\documentclass[sigconf]{acmart}
\usepackage{booktabs} 
\DeclareMathOperator*{\argmin}{argmin}

\usepackage{flushend}
\usepackage{algpseudocode}
\usepackage{algorithm}
\usepackage{algorithmicx}

\usepackage{array}
\newcolumntype{L}[1]{>{\raggedright\let\newline\\\arraybackslash\hspace{0pt}}m{#1}}
\newcolumntype{C}[1]{>{\centering\let\newline\\\arraybackslash\hspace{0pt}}m{#1}}
\newcolumntype{R}[1]{>{\raggedleft\let\newline\\\arraybackslash\hspace{0pt}}m{#1}}

\begin{document}
\title{Integrated Face Analytics Networks through \\ Cross-Dataset Hybrid Training}

\author{Jianshu Li $^{1,4}$\quad Shengtao Xiao $^2$ \quad Fang Zhao $^2$ \quad Jian Zhao $^2$ \quad Jianan Li $^3$ }
\author{Jiashi Feng $^2$ \quad Shuicheng Yan $^2$ \quad Terence Sim $^1$ }

\affiliation{%
  \institution{$^1$ School of Computing, National University of Singapore, Singapore}
  \institution{$^2$ Electrical \& Computer Engineering, National University of Singapore, Singapore}
  \institution{$^3$ Beijing Institute of Technology University, P. R. China}
  \institution{$^4$ SAP Innovation Center Network Singapore, Singapore}
}
\email{{jianshu, xiao_shengtao, zhaojian90}@u.nus.edu, lijianan15@gmail.com}
\email{{elezhf, elefjia, eleyans}@nus.edu.sg, tsim@comp.nus.edu.sg}

\renewcommand{\shortauthors}{J. Li et al.}

\copyrightyear{2017}
\acmYear{2017}
\setcopyright{acmlicensed}
\acmConference{MM'17}{}{October 23--27, 2017, Mountain View, CA, USA.}
\acmPrice{15.00}
\acmDOI{https://doi.org/10.1145/3123266.3123438}
\acmISBN{ISBN 978-1-4503-4906-2/17/10}

\fancyhead{}
\settopmatter{printacmref=false, printfolios=false}

\begin{abstract}
Face analytics benefits many multimedia applications. It consists of a number of tasks, such as facial emotion recognition and face parsing, and most existing approaches generally treat these tasks independently, which limits their deployment in real scenarios. In this paper we propose an integrated Face Analytics Network (iFAN), which is able to perform multiple tasks jointly for face analytics with a novel carefully designed network architecture to fully facilitate the informative interaction among different tasks. The proposed integrated network explicitly models the interactions between tasks so that the correlations between tasks can be fully exploited for performance boost. In addition, to solve the bottleneck of the absence of datasets with comprehensive training data for various tasks, we propose a novel cross-dataset hybrid training strategy. It allows ``plug-in and play'' of multiple datasets annotated for different tasks without the requirement of a fully labeled common dataset for all the tasks. We experimentally show that the proposed iFAN achieves state-of-the-art performance on multiple face analytics tasks using a single integrated model. Specifically, iFAN achieves an overall F-score of $91.15\%$ on the Helen dataset for face parsing, a normalized mean error of $5.81\%$ on the MTFL dataset for facial landmark localization and an accuracy of $45.73\%$ on the BNU dataset for emotion recognition with a single model. 
\end{abstract}

\keywords{Integrated network; Face analytics; Cross-dataset hybrid training; Feedback loop; Task interaction}

\maketitle

\begin{figure}
\includegraphics[width=\linewidth]{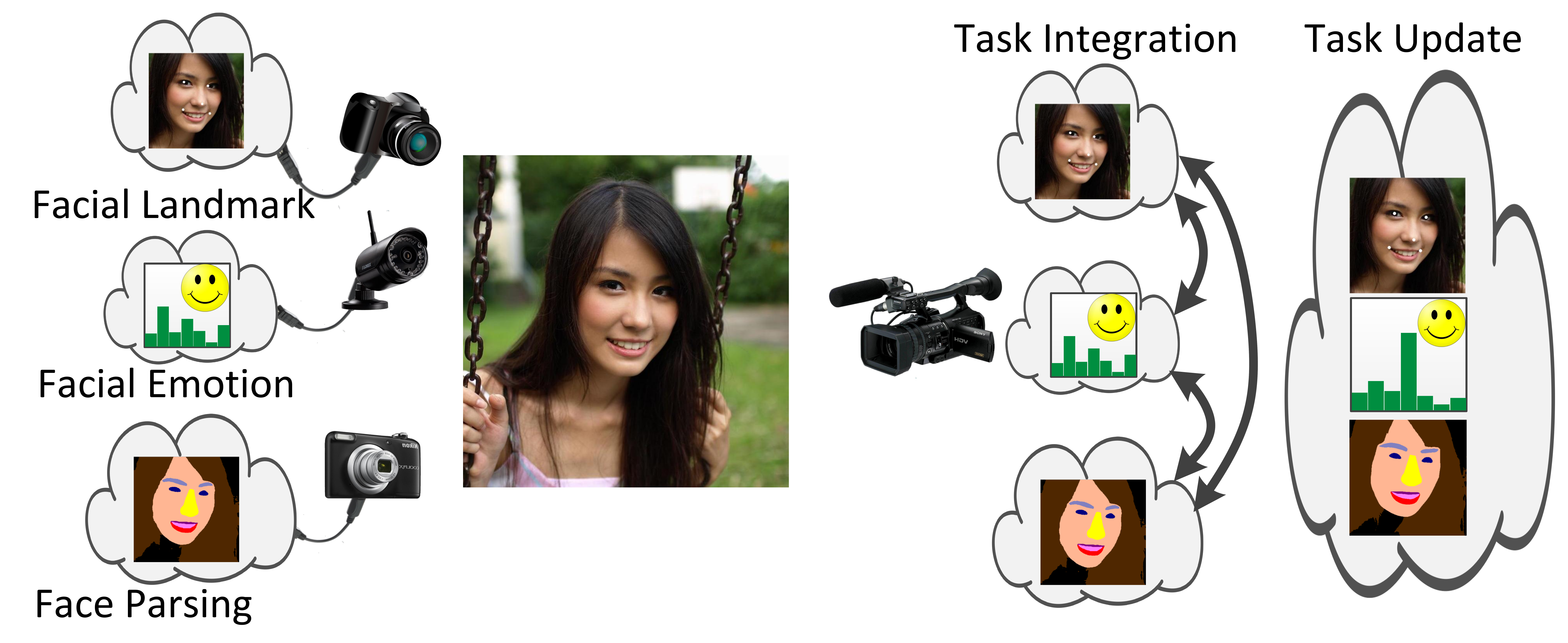}
\caption{Our motivation. Traditionally different face analytics tasks are performed by different models (symbolized by cameras). Each model aims at a specific task. In contrast, iFAN solves all tasks by an integrated  model, which exploits the correlations among the tasks to enable full task interaction and performance boost, serving as a one-stop solution to all the face analytics problems of interest. }\label{fig:motivation}
\end{figure}

\section{Introduction}
Face analytics is essential for human-centric multimedia research and applications. Face analytics tasks include face detection~\cite{chen2016supervised}, facial landmark localization~\cite{xiao2016robust,zhang2014facial}, face attribute prediction~\cite{liu2015faceattributes}, face parsing~\cite{smith2013exemplar,zhou2015interlinked}, facial emotion recognition~\cite{dhall2016emotiw,li2016happiness}, face recognition~\cite{guo2016ms,li2016robust},~\emph{etc}. 

Traditionally, different face analytics tasks are treated separately and performed by designing different models. But in some scenarios, people need to address multiple face analytics tasks. For example, for facial emotion recognition task, people also need to address facial landmark localization task as the input to facial emotion recognition task needs to be aligned by the detected facial landmarks. So it is attractive  to design an integrated face analytics network which performs multiple tasks in one go.

In this work we propose an integrated face analytics network (named iFAN).  Different from existing approaches where separate models are used for different tasks, iFAN is a powerful model to solve different tasks simultaneously, enabling full task interactions within the model. See Figure~\ref{fig:motivation}. In additon, the iFAN uses a novel cross-dataset hybrid training strategy to effectively learn from multiple data sources with orthogonal annotations, which solves the bottleneck of lacking complete training data for all involved tasks.  

The proposed iFAN uses a carefully designed network architecture that allows for informative interaction between tasks. It consists of four components: a shareable feature encoder, feature decoders, feature re-encoders and a task integrator. The shareable feature encoder, which is the backbone network, learns rich facial features that are discriminative for different tasks. Each of the feature decoders produces the prediction on top of the learned features for one specific task. To promote interactions among different tasks within iFAN, the feature re-encoders and task integrator are introduced. The feature re-encoders in iFAN  transform the task specific predictions back to feature spaces. We use the term ``re-encoder'' to stress the function of converting the predictions back to the feature space. Specifically the feature re-encoders take as input raw predictions and generate encoded features of the predictions. The feature re-encoders can align the features for different tasks to similar semantic levels to facilitate the task interaction process. Based on the representations from re-encoders, the task integrator in iFAN integrates the encoded predictions of different tasks into multi-resolution and multi-context features that facilitate the inter-task interactions. Specifically, with access to the encoded predictions of all tasks, the task integrator provides the full context information for the task interactions. 
It introduces a feedback loop, which connects the integrated context information back to the backbone network,
which is beneficial for performing multiple tasks simultaneously. 

To the end of jointly addressing different tasks, one bottleneck is the absence of datasets with complete training data for all the tasks of interest. Usually each dataset only provides annotations for a specific task (\emph{e.g.} emotion category for emotion recognition, segmentation mask for face parsing), and it is very hard to find a dataset with a complete set of labels for all the tasks of interest. Thus we propose a new cross-dataset hybrid training strategy to enable iFAN to learn from multiple data sources and perform well on all tasks simultaneously. 
The proposed cross-dataset hybrid training strategy can effectively model the statistical differences across different datasets to reduce the negative impacts of such differences. With the proposed training strategy, the iFAN does not require complete annotations for all the tasks over a single dataset. Instead, this training strategy allows iFAN to learn from multiple data sources without annotation overlapping. Such ``plug-in and play'' feature greatly increases the flexibility of iFAN.

The iFAN uses only one network for multiple face analytics tasks, enabling users to customize their own combination of tasks for iFAN to  perform simultaneously. The model size, computation complexity and inference time are linearly reduced compared with separate models. Moreover, iFAN goes a step further to analyze the correlations between the tasks, which enables interaction with each other for performance boost. 

It is worth noting that iFAN is different from multi-task learning. Unlike the simple parameter sharing scheme in commonly used multi-task learning models, iFAN explicitly models the interaction between different tasks. More than merely sharing a common feature space, the outputs from different tasks also jointly influence the predictions of other tasks. 
Besides, the proposed iFAN is able to learn from multiple data sources with no overlapping, where traditional multi-task learning approaches will fail.  Thus the expensive cost of collecting comprehensive training data for all involved tasks can be substantially reduced. Our work is also different from transfer learning which considers to learn the same task from different datasets. In contrast, our proposed cross-dataset hybrid learning is able to utilize the useful knowledge on learning different tasks from non-overlapping datasets.

\section{Related Work}
In this section, we briefly review related work, including standard multi-task deep learning and specific face analytics. 

\paragraph{Multi-Task Deep Learning} Deep neural network has outstanding learning capacity and thus it is possible for it to learn to perform multiple tasks at the same time. For example, in the scenario of image analysis, the features learned by deep neural networks at bottom  layers are known to characterize low-level features such as edges and blobs, which are common for all image analysis tasks so they are universal for different vision tasks. Some work shows that the higher level features can also be shared across different tasks. For instance, Fast RCNN~\cite{girshick2015fast} uses the same network to perform object confidence score prediction and bounding box regression. In addition to these two tasks, Faster RCNN~\cite{ren2015faster} uses the same network to generate region proposals as well. A recent work Mask RCNN~\cite{he2017mask} adds a segmentation task,~\emph{i.e.} mask prediction, to the same trunk of the network. TCDCN~\cite{zhang2014facial} uses a deep network to perform the task facial landmark localization and face attribute prediction (such as facial emotion, pose) and shows that adding face attribute prediction can help improve the performance of facial landmark localization. MTCNN~\cite{zhang2016joint} performs the task of face detection and facial landmark localization together and HyperFace~\cite{ranjan2016hyperface} performs face detection, landmark localization, pose estimation and gender recognition in one network. We can see that a single network is capable of performing multiple tasks together. However, the informative relations among different tasks are not explored in these previous works. Existing multi-task learning networks generally focus on learning  common representations for different tasks. All the tasks are learned in parallel and the useful feedback information from one task for other tasks is not modeled. A recent work~\cite{bilen2016integrated} models task interactions with integrated perception, but only simple hand-crafted prediction encoding scheme is used. In contrast to existing multi-task learning models, our proposed iFAN explicitly models the interaction between different tasks with learnable feature re-encoders, and the feedback information effectively contributes to the representation learning as well as boosting performance for all the tasks. 

\paragraph{Face Analytics}  A lot of research has been conducted on individual face analytics, especially on analyzing challenging unconstrained faces, \emph{i.e.} faces in the wild. The field of face analytics has been accelerated by emergence of large scale unconstrained face datasets. One of the large face attribute prediction datasets, CelebA, is proposed in~\cite{liu2015faceattributes}. MsCeleb-1M dataset~\cite{guo2016ms} is a big face-in-the-wild dataset for face recognition. Most of the datasets focus on one task with labels only for that task. There are some datasets which have multiple sets of labels for different tasks. Annotated Facial Landmarks in the Wild (AFLW)~\cite{tugraz:icg:lrs:koestinger11b} provides a large-scale collection of annotated face images with face location, gender and $21$ facial landmarks.  Multi-Task Facial Landmark (MTFL) dataset~\cite{zhang2014facial} contains annotations of five facial landmarks and attributes of gender, head pose,~\emph{etc}. However, such datasets can only cover a subset of all the face analytics tasks. Thus it is usually not easy to find a dataset with a complete label set for combinations of tailored tasks of interest. Thus a model which allows ``plug-in and play'' of multiple datasets from different sources is of great practical value but is still absent. 

\begin{figure*}
\includegraphics[width=0.9\linewidth]{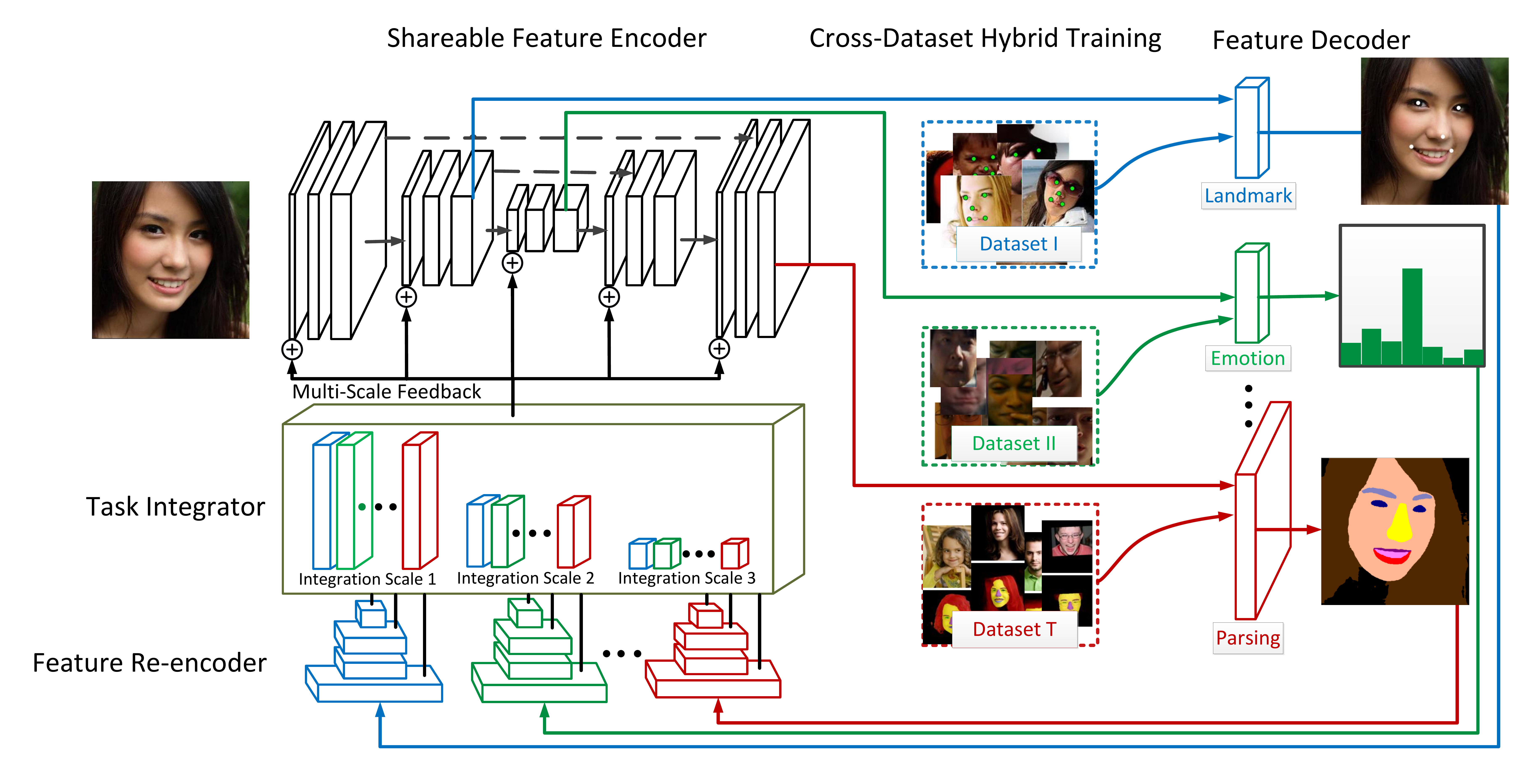}
\caption{Overall structure of iFAN. Black blocks denote the backbone network for learning shareable features. Each colored block is associated with one task, namely, blue for facial landmark localization, green for facial emotion recognition and red for face parsing. Each task has its own feature decoder and feature re-encoder, which perform task prediction and prediction encoding,  respectively. The task integrator integrates the encoded prediction features from different tasks in multiple scales of spatial resolutions. The integrated multi-resolution features are then fed back into the respective multiple feature spaces in the backbone network. Different tasks are associated with different training datasets without overlapping in both images and annotations. The whole network is trained with the proposed cross-dataset hybrid training strategy.}
\label{fig:structure}
\end{figure*}

\section{Proposed Method}
In this section, we elaborate on the proposed integrated Face Analytics Network (iFAN). Its overall structure is shown in Figure~\ref{fig:structure}. The backbone network of iFAN learns shareable features for different face analytics tasks, and different tasks take in features from different layers within the backbone network to perform prediction. In Figure~\ref{fig:structure}, three tasks are illustrated, including facial landmark localization, facial emotion recognition and face parsing, each of which employs a feature decoder to make predictions for the corresponding task. Different from existing multi-task learning models, iFAN introduces task-specific feature re-encoders to facilitate task interaction. The feature re-encoders takes predictions from different tasks and re-encode them back to semantically rich feature spaces across the tasks in multiple spatial resolutions. iFAN also has a task integrator, which aggregates the re-encoded features from different tasks and feeds them back to the backbone network for task interaction and improve the shareable feature learning. To solve the data incompleteness problem, we propose a novel cross-dataset hybrid training strategy, which allows iFAN to effectively learn from multiple datasets with orthogonal annotations, without requiring any dataset with comprehensive annotations.

\subsection{Preliminary}
We first introduce the problem setup formally. Suppose there are $T$ tasks under consideration and there is a training dataset with a complete set of labels for all the $T$ tasks: $D=\{(x_i, y_i^1, y_i^2, \cdots, y_i^T)_{i=1}^{N}\},$ where $x_i$ is the $i$-th data sample and $y_i^t, \forall t=1, 2, \cdots, T$ is the corresponding label for the $t$-th task. The traditional multi-task learning problem seeks to find the set of parameters such that
\begin{equation}\label{eqn:mtl_full_label}
(\hat{\theta}^S, \hat{\theta}^1, \cdots, \hat{\theta}^T) =\argmin_{\theta^S, \theta^1, \cdots, \theta^T}
\sum_{t=1}^T \frac{1}{N}\sum_{i=1}^{N}\ell \left(f_{\theta^t}\circ f_{\theta^S}(x_i),y_i^t \right),
\end{equation}
where $\ell$ denotes the loss between the prediction and the ground truth label, $\theta^S$ is the shared network parameter and $\theta^t$ is the parameter to perform the $t$-th task. 
Although widely used, the multi-task learning in Eqn.~\eqref{eqn:mtl_full_label} can be improved from two perspectives. First, the formulation only implicitly models the interactions between tasks through the shared data feature and an explicit modeling is not present. Second, the model requires a dataset with complete labels for all tasks, which is rather difficult to collect. It is beneficial if we can get rid of this requirement. We propose to make these two improvements over the original multi-task learning through a new integrated network model and a new cross-dataset learning, detailed in the following two subsections.

\subsection{Task Integrator}
In the traditional multi-task learning formulated in  Eqn~\eqref{eqn:mtl_full_label}, different tasks share common features for exploiting correlations among different tasks. However, the interactions among different tasks are not explicitly modeled\textemdash they only interact with each other through error back-propagation to contribute to the learned feature and such implicit interactions are not  controllable. The prediction of a certain task is certainly  benefited from   other related tasks for face analytics, but this dependency is rarely modelled in the traditional multi-task learning. The proposed iFAN  explicitly models and exploits beneficial feedback from different  tasks through  a task integrator. The task integrator integrates the features from the predictions of all the tasks, and feeds them back to the backbone network. In this way the task integrator provides the information of other tasks' predictions in order to further refine the prediction of the current task under consideration. 

As the predictions are decoded by different task-specific decoders, the predictions of different tasks lie in different semantic spaces and it is not trivial to properly model the inter-task interactions. We propose to use the task-wise feature re-encoder to encode the predictions from different tasks into a set of semantically rich features. The re-encoded features from different tasks are integrated by the task integrator, and then fed-back to the backbone network. As different tasks draw features from different layers in the backbone network, we feedback the re-encoded features to multiple layers in the backbone network with different spatial sizes. The feature re-encoder naturally generates a pyramid of features with different spatial sizes, and all of them are used in the multi-layer, multi-resolution feedback. The encoded features facilitate interactions among different tasks during training and deploying the integrated face analytics model.

The proposed iFAN uses a task integrator and task-specific feature re-encoders to explicitly model task interactions. Formally the task integrator models the effects of other tasks by creating a set of integrated feature spaces where the predictions from different tasks are encoded to
\begin{equation}
\label{eqn:int_def}
f_{\rm{INT}}(x) =\sum_{t=1}^T f_{\theta_e^{t}}\circ f_{\theta^{t}} \circ f_{\theta^S}(x)+f_{\theta^S}(x), 
\end{equation}
where $f_{\theta^S}(x) $ is the learned feature shareable across multiple tasks for one input sample $x$ and ${f_{\theta^{t}}\circ f_{\theta^S}(x)}$ is the prediction of the $t$-th task based on $f_{\theta^S}(x) $. Parametrized by $\theta_e^t$, the feature re-encoder of the $t$-th task performs encoding of  the predictions of the $t$-th task, as represented by ${f_{\theta_e^{t}}\circ f_{\theta^{t}} \circ f_{\theta^S}(x)}$.  The summation here denotes feature level integration. This encoding space of an input sample $x$ aggregates the features from not only the original feature, but also the encoded predictions from all the tasks. 

Based on $f_{\rm{INT}}(x)$, we can reformulate Eqn.~\eqref{eqn:mtl_full_label} as
\begin{equation}\label{eqn:mtl_full_label_interaction}
\begin{split}
(\hat{\theta}^S, \hat{\theta}^1, \cdots, \hat{\theta}^T, \hat{\theta}_e^1, \cdots, \hat{\theta}_e^T) &= \\
\argmin_{\theta^S, \theta^1, \cdots, \theta^T, \theta_e^1, \cdots, \theta_e^T}
&\sum_{t=1}^T \frac{1}{N}\sum_{i=1}^{N}\ell \left(f_{\theta^t}\circ f_{\rm{INT}}(x_i),y_i^t \right).
\end{split}
\end{equation} 

We can see that the prediction of the $t$-th task is made from  the integrated feature space $f_{\rm{INT}}$, which contains features from all the tasks. The integrated feature space provides rich information and context cues for the predictions of the $t$-th task. 

The formulation in Eqn.~\eqref{eqn:int_def} extends naturally to an iterative updating formulation:
\begin{equation}
\label{eqn:int_iter_def}
f_{\rm{INT}_{I}}(x)=
\begin{cases}
\sum_{t=1}^T f_{\theta_e^{t}}\circ f_{\theta^{t}} \circ f_{\rm{INT}_{i-1}}(x)+f_{\theta^S}(x), &\text{for } I>1 \\
f_{\theta^S}(x), & \text{for } I = 0.
\end{cases}
\end{equation}

With this iterative formulation, Eqn.~\eqref{eqn:mtl_full_label_interaction} becomes 
\begin{equation}\label{eqn:mtl_full_label_interaction_iter}
\begin{split}
(\hat{\theta}^S, \hat{\theta}^1, \cdots, \hat{\theta}^T, \hat{\theta}_e^1, \cdots, \hat{\theta}_e^T) &= \\
\argmin_{\theta^S, \theta^1, \cdots, \theta^T, \theta_e^1, \cdots, \theta_e^T}
\sum_{t=1}^T &\frac{1}{N}\sum_{i=1}^{N}\sum_{I=0}^{\rm{ITER}}\ell \left(f_{\theta^t}\circ f_{\rm{INT}_I}(x_i),y_i^t \right),
\end{split}
\end{equation}
where $\rm{ITER}$ is the maximal iteration of task interactions. When $\rm{ITER}=0$, Eqn.~(\ref{eqn:mtl_full_label_interaction_iter}) reduces to ordinary multi-task learning formulation in Eqn.~\eqref{eqn:mtl_full_label}. With $\rm{ITER} > 0$, the iterative refinement is turned on with the feedback loop (the connection from the task integrator to the backbone network in Figure~\ref{fig:structure}). With the feedback loop and  the iterative refinement process, the task integrator enables interactions of different tasks and helps make better predictions.

\subsection{Cross-dataset Hybrid Training}
Based on Eqn.~\eqref{eqn:mtl_full_label_interaction_iter}, we propose a cross-dataset hybrid training strategy to bypass the requirement of data fully labeled for all the $T$ tasks, as it is difficult to satisfy in real scenarios. We consider the more realistic cases where data annotations are incomplete and aim at an integrated network model for all the $T$ tasks with incomplete training information. Each task is provided with a specific training dataset which is denoted as $D_t=\{(x_i^t, y_i^t)_{i=1}^{n_t}\},$ where $x_i^t$ is the $i$-th input data point for the $t$-th task, and $y_i^t$ is the corresponding label. There is no overlapping between datasets for different tasks, \emph{i.e.} $x_i^t \neq x_j^{t'}, \forall i,j,t,t',\and t' \neq t$. This setting is quite common in reality. A trivial and straightforward solution is to train $T$ models for the $T$ tasks, each with the respective training data $D_t$. Such a trivial solution clearly leaves the relations between tasks un-modeled and thus is sub-optimal. In the proposed iFAN, we build an integrated network, which is trained on multiple data sources, yet still enjoys the benefits of multi-task learning.

When training from multiple data sources $D_t, \hspace{1mm}t=1,2,\cdots, T$, we cannot optimize the parameters for all the tasks as in Eqn.~\eqref{eqn:mtl_full_label_interaction_iter}, but need to focus on one of the tasks every time. When we optimize the integrated network for the $t$-th task, we have 
\begin{equation}\label{eqn:mtl_full_label_hybrid}
\begin{split}
(\hat{\theta}^S, \hat{\theta}^t,\hat{\theta}_e^1, \cdots, \hat{\theta}_e^T) &= \\
\argmin_{\theta^S, \theta^t, \theta_e^1, \cdots, \theta_e^T}
&\frac{1}{n_t}\sum_{i=1}^{n_t} \sum_{I=0}^{\rm{ITER}}\ell \left(f_{\theta^t}\circ f_{\rm{INT}_I}(x_i^t),y_i^t \right).
\end{split}
\end{equation}
Here, we only use the supervision information from the $t$-th task, but the integrated feature $f_{\rm{INT}_I}(x_i^t)$ incorporates the prediction information from all other tasks for the input sample $x_i^t$ in the $t$-th task. Optimizing Eqn.~\eqref{eqn:mtl_full_label_hybrid} directly will lead only to the optimal solution to the $t$-th task, making the common feature space $\theta^S$ bias towards the $t$-th task. Such a situation is undesired and our final target is an optimal  solution to all the tasks. 

In iFAN, we use a strategic alternative training scheme to achieve the cross-dataset hybrid training. We use $\Delta^t_I(\cdot)$ to denote the operation of one gradient update of the involved  parameters with the provided data $(\cdot)$ in the $I$-th task interaction towards the direction of optimizing Eqn.~\eqref{eqn:mtl_full_label_hybrid} for the $t$-th task. Then the cross-dataset hybrid training strategy can be summarized in Algorithm~\ref{algo:1}.

\begin{algorithm}[b]
  \caption{Cross-dataset Hybrid Training Strategy
    \label{algo:1}}
  \begin{algorithmic}[1]
    \Require{Randomly initialized $\theta^S, \theta^1, \theta^2, \cdots, \theta^T, \theta_e^1, \theta_e^2,\cdots, \theta_e^T$}, Training data $D_t=\{(x_i^t, y_i^t)_{i=1}^{n_t}\}$, Batch size of the gradient descent $n_b$, Total number of task interaction iterations ITER, Number of Pre-training epochs $E_P$
    \Statex
      \For{$t \gets 1 \textrm{ to } T$}
      \For{$e \gets 1 \textrm{ to } E_P$}
	\While{$D_t$ is not traversed }      
	 \State Sample $n_b$ data points from $D_t$ as $\{(x_i^t, y_i^t)_{i=1}^{n_b}\}$   
      \State $\theta^S, \theta^t \gets \Delta^t_0(\{(x_i^t, y_i^t)_{i=1}^{n_b}\})$  
     \EndWhile
      \EndFor 
    \EndFor
      \While{$\theta^S, \theta^1, \cdots, \theta^T, \theta_e^1, \cdots, \theta_e^T$ are not converged}
      \For{$t \gets 1 \textrm{ to } T$}
        \State Sample $n_b$ data points from $D_t$ as $\{(x_i^t, y_i^t)_{i=1}^{n_b}\}$   
            \For{$I \gets 1 \textrm{ to } \rm{ITER}$}
   \State $\theta^S, \theta^t, \theta_e^1, \cdots, \theta_e^T \gets \Delta^t_I(\{(x_i^t, y_i^t)_{i=1}^{n_b}\})$
	       \EndFor   
      \EndFor
      \EndWhile
  \end{algorithmic}
\end{algorithm}

The cross-dataset hybrid training contains two stages: task-wise pre-training and batch-wise fine-tuning. For the task-wise pre-training, we loop through every dataset to learn the common features and the task specific feature decoders so that task specific feature decoders have the ability to perform the task. During the process, the common feature may bias towards the latest task, to which the batch-wise fine-tuning is used as a complement. The feature re-encoders and task integrator are also added in the second stage so that the task interactions are enabled.  Since with pre-training, each feature decoder can make reasonable predictions about its own task, we turn on task interaction only in the second stage. In the second stage, each task will take turns to update its parameters  with the guidance of its label information. Moreover, each task has an equal number of training samples from its training set for each update. It addresses the issue of imbalanced numbers of training samples from multiple datasets, and the resultant network will not bias towards any of the training sets with larger numbers of training data.

Empirically, we find that task-dependent batch normalization parameters are important in the backbone network, which agrees with~\cite{bilen2017universal}. Different datasets vary in terms of statistical distributions such as image quality, illumination condition on faces,~\emph{etc}. The task-wise batch normalization will effectively address the shifts of statistical distributions of the features across different datasets to facilitate the learning of useful and robust common features within multiple datasets. Although simple, we experimentally demonstrate that together with the task integrator, the cross-dataset hybrid training strategy effectively helps the integrated face network learn from multiple data sources.

\section{Experiments}
We conduct experiments to validate the power of iFAN with multiple face tasks, and also provide ablation study in this section.

\subsection{Experimental Setting} 
\subsubsection{Datasets} 
In the experiments, we consider three important fine-grained face analytics tasks including face parsing, facial landmark localization, and facial emotion recognition. Each task is associated with a different dataset.

The task of face parsing (or face segmentation, face labeling) aims to predict semantic categories for all pixels in face images. We use the popular Helen dataset~\cite{le2012interactive} for this task. It contains $2{,}330$ images with accurate and detailed annotations of the primary facial components. The work~\cite{smith2013exemplar} modifies the original Helen dataset to suit a face parsing task by generating segmentation masks for the facial components (such as eyes, nose, mouth, \emph{etc.}) and hair regions. The categories in the Helen dataset include eyes, eyebrows, nose, inside mouth, upper lip, lower lip, face skin and hair. Every pixel needs to be classified into one of these categories or background. 

Facial landmark localization aims to find coordinates of pre-defined facial landmarks. For this task, we use Multi-Task Facial Landmark (MTFL) dataset~\cite{zhang2014facial}. It contains $12{,}995$ face images annotated with $5$ facial landmarks, namely, eye centers, nose tip and mouth corners. The images in the dataset contain various pose angles and occlusion, thus it is challenging to accurately localize facial landmarks.

For facial emotion recognition, we use BNU Large-scale Spontaneous Visual Expression Database (BNU-LSVED)~\cite{sun2015multi, sun2016bnu}. It is designed to capture facial emotions in the educational environment. It contains $1{,}572$ subjects, with totally about $63{,}000$ images and $7$ emotions: ``Happy'', ``Surprised'', ``Disgusted'', ``Puzzled'', ``Concentrated'', ``Tired'' and ``Distracted''. The original dataset contains images in videos and there are a lot of near duplicates. We adopt this dataset for the task of static emotion recognition by sampling images from the video sequences. The resultant dataset  after sampling contains about $6{,}100$ images. 

\emph{Different tasks have different sets of labels and there is no overlap between them}. Currently, there is no dataset that covers every possible combination of face analytics tasks of interest. Our proposed iFAN model and the cross-dataset hybrid training strategy allow any task to be plugged into the integrated framework without worrying the statistical differences among the different datasets.

\subsubsection{Implementation Details}
In iFAN, we use fully convolutional DenseNets~\cite{jegou2016one} as the backbone network, considering its outstanding ability at re-using features learned at different layers. The fully convolutional DenseNet has a down-sampling stage and an up-sampling stage. In both stages, we use $5$ dense blocks with $3$ layers in each block and a growth rate of $12$.  All the convolutional layers in the dense blocks are resolution preserving with stride $1$ and kernel size $3$, except for the initial convolution where we use kernel size $7$ to increase the receptive field. At the end of each dense block in the down-sampling stage, we use average pooling to halve the spatial dimension. At the end of each dense block in the up-sampling stage, we use sub-pixel sampling layer~\cite{shi2016real} followed by a convolutional layer to double the feature spatial dimension. The input size of each face is $128 \time 128$. In the down-sampling stage, the spatial resolution of the feature maps reduce from $128$ to $64$, $32$, $16$, $8$ and $4$ after each average pooling operation. Inversely, in the up-sampling stage, the spatial resolution of the features gradually increases from $4$ back to $128$. 

For facial landmark localization, the features with dimension $8\times 8$ in the down-sampling stage are used as input for the landmark decoder which performs a regression to the normalized coordinates of the facial landmarks with the Euclidean distance loss. For the face parsing task, we use the features with dimension $128$ at the end of the up-sampling stage as input to the face parsing decoder which performs a per-pixel prediction of the pixel label with a categorical cross entropy loss. For the facial emotion recognition task, we use the feature with spatial size $4$ as input for the attribute decoder which performs a single prediction of the attribute label with a categorical cross entry loss. Note that for the face parsing task, the loss is calculated on the $128\times 128$ prediction map. But the prediction is done by resizing the prediction map to the original size of the input with bilinear interpolation and then comparing with the ground truth label for each pixel. 

For the feature re-encoders, we design different encoders for different tasks. For the facial landmark localization task, we construct $128\times 128$ point heat maps with hot values indicating the locations of the landmarks. We enlarge the one-hot point heat map to a radius of $5$ pixels. Then the point heat maps are used as inputs into alternating convolution layers and max pooling layers to perform feature encoding of the landmark predictions. For the face parsing task, we feed the parsing prediction map, which also has the size of $128 \times 128$ and contains cues for face parsing results, into the feature re-encoder with alternating convolution layers and max pooling layers. For the attribute prediction task, we use several fully connected layers to encode the predicted probability vectors, and tile the encoded feature to the corresponding spatial dimensions. The feature re-encoders convert the raw predictions of different tasks into a pyramid of semantically-rich features to facilitate task interaction and integration. The integration in Eqn.~\eqref{eqn:int_iter_def} is realized by feature concatenation.

For training, we use mini-batch gradient descent with batch size $24$, $64$ and $96$ for parsing, landmark and emotion, respectively. The optimizer used is RMSprop~\cite{tielemanrmsprop}. For pre-training, each task is trained with learning rate $10^{-3}$ for $30$ epochs. For fine-tuning, the total number of training epochs is $200$ and the learning rate reduces from $10^{-3}$ to $10^{-6}$ during the entire training process.

\subsubsection{Evaluation Metrics}
\paragraph{Face parsing}
For face parsing we follow~\cite{smith2013exemplar} and use F-score for evaluation, which is the harmonic mean of precision and recall, to measure the performance. We report the F-score for all the classes in the Helen dataset, as well as two additional scores for all the components associated with mouth (Month-All) and overall score to keep the comparison consistent with~\cite{smith2013exemplar} and~\cite{liu2015multi}. 

\paragraph{Facial Landmark Localization}
For facial landmark localization, we report the results on two widely used metrics~\cite{zhang2014facial, xiao2016robust}, \emph{i.e.} normalized mean error and failure rate. The normalized mean error is the distance between the estimated landmark and the ground truth, normalized with respect to the inter-ocular distance. A failure happens when the normalized mean error is larger than $10\%$.
\paragraph{Facial Emotion Recognition} For facial emotion recognition, we adopt the accuracy of the prediction as compared with the ground truth annotations as the evaluation metric. 

\begin{figure}

\includegraphics[width=\linewidth]{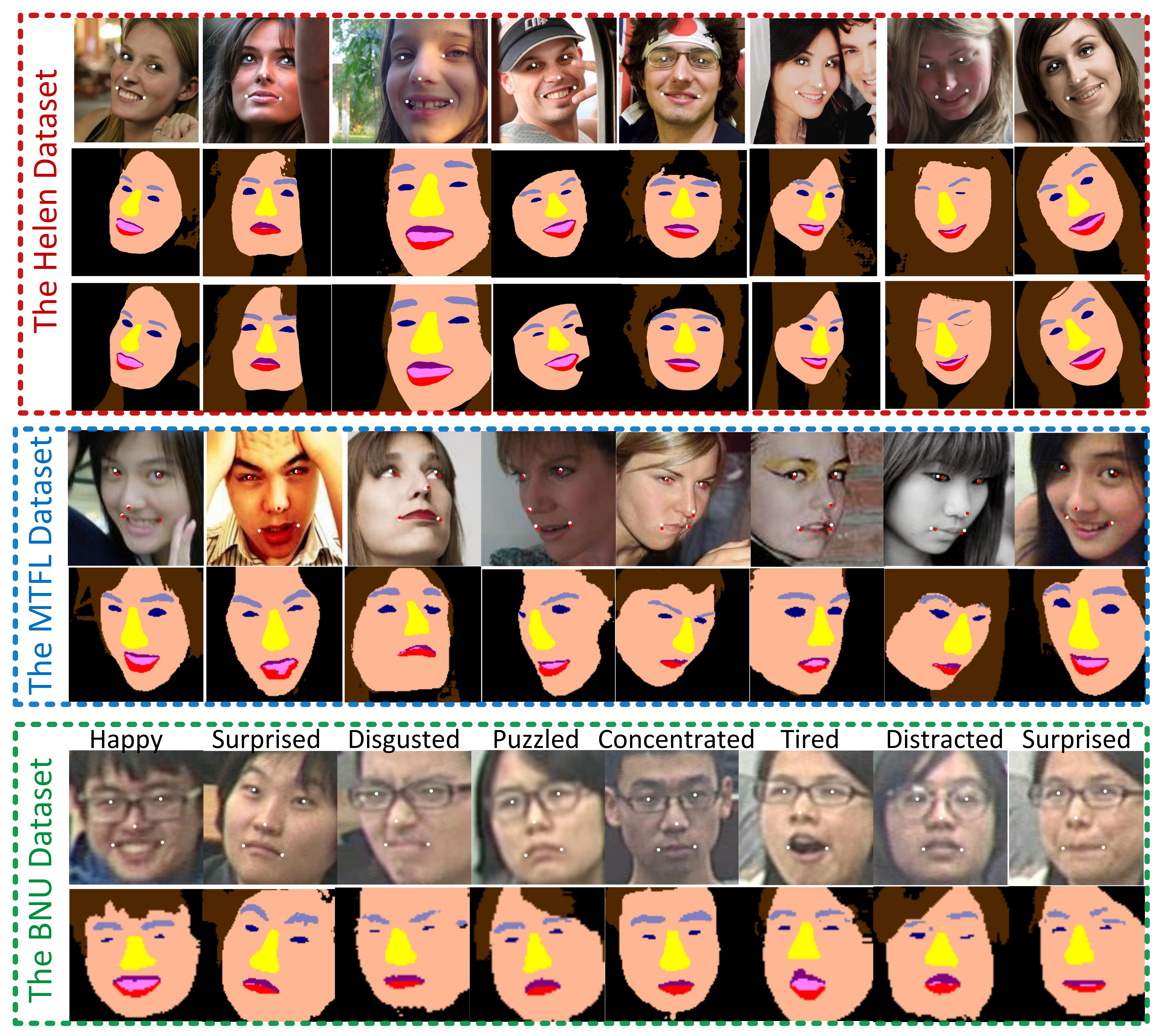}

\caption{Some results of face analytics from iFAN. The top block shows images from the Helen dataset (which is designed for face parsing). White dots on the faces indicate detected facial landmarks. The second row shows the predicted face parsing results and the third row is the parsing ground truth. The second block in blue shows the face images from the MTFL dataset (designed for landmark detection). White dots are detected landmarks and red ones are given ground truth.  The second row in this block shows the face parsing results. The bottom block shows face images from the BNU dataset with detected lardmarks, face parsing maps and correctly predicted facial emotions. All the results demonstrate iFAN is good at modeling interactions between multiple tasks, even there is no complete training image set. The figure is best viewed in color. }\label{fig:quality}
\end{figure}

\subsection{Results and Comparison}
We compare the performance of the proposed iFAN with well established baseline methods. We consider two multi-task settings for iFAN: 1) performing facial landmark localization and face parsing simultaneously (denoted as 2T); 2) performing facial landmark localization, face parsing and emotion recognition simultaneously (denoted as 3T). We report the performance of iFAN  and state-of-the-art baseline methods. For facial landmark localization, we compare with  state--of-the-art TSPM~\cite{zhu2012face}, ESR~\cite{cao2014face}, CDM~\cite{yu2013pose}, RCPR~\cite{burgos2013robust}, SDM~\cite{xiong2013supervised}, TCDCN~\cite{zhang2014facial} and MTCNN~\cite{zhang2016joint}. For face parsing, we compare with Generative Shape Regularization Model (GSRM)~\cite{gu2008generative}, Examplar~\cite{smith2013exemplar}, Multi-Objective~\cite{liu2015multi} and iCNN~\cite{zhou2015interlinked}. For our results, we follow the official training/testing split of the MTFL dataset in~\cite{zhang2014facial} and the Helen dataset as described in~\cite{smith2013exemplar}, and report the performance on the respective testing set. The second setting involves BNU-LSVED, which is a relatively new one without public training/testing split protocols, we choose $20\%$ subjects in each emotion category as the testing set and the rest are used for training/validation (with no overlapping subjects in training and testing sets). We use the same network structure to train different strong baselines for comparison.  No other external datasets are used during the training process for both the two settings.

\begin{figure}[t]
\includegraphics[width=\linewidth]{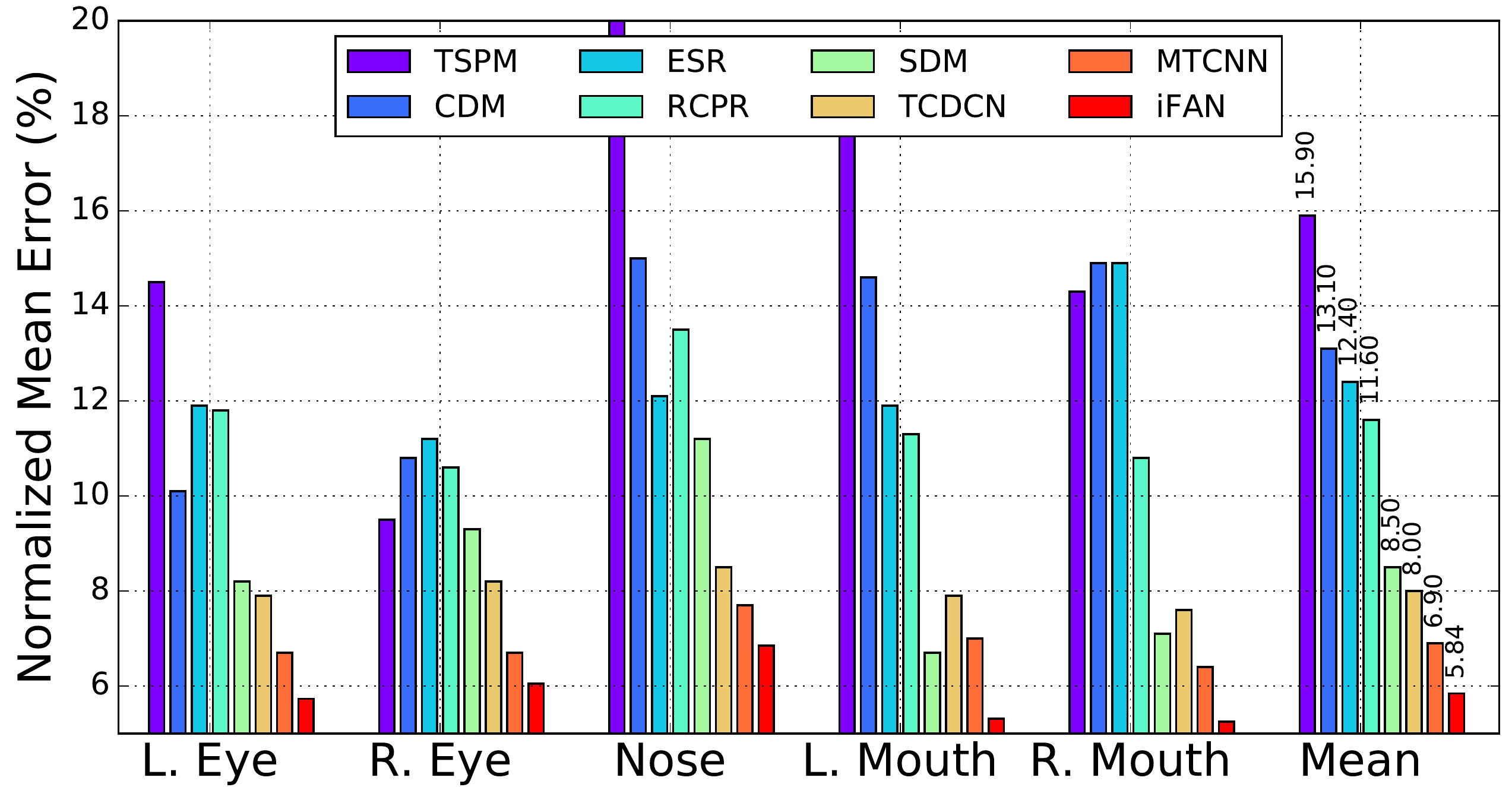}

\caption{Normalized mean errors of different methods for different landmarks.  The values for TCDCN are the best results achieved in~\cite{zhang2014facial}. The results for other baselines are obtained from~\cite{zhang2016joint}. L. Eye denotes left eye center and R. Mouth means right mouth corner. }\label{fig:pts}
\end{figure}

\subsubsection{Facial Landmark Localization}
The performance on the facial landmark localization task with iFAN and other baselines is shown in Figure~\ref{fig:pts}. The normalized mean errors on different landmarks for different methods are illustrated. iFAN achieves the best performance for all the landmarks, outperforming state-of-the-art performance reported before. Specifically, the NMEs for both the two-task (2T) and three-task (3T) settings and the performance over different iterations of interactions (Iter0, Iter1 and Iter2) are detailed in Table~\ref{tab:pts}. For Iter0, there is no interaction between the tasks, and  iFAN reduces to an ordinary multi-task learning network, except for it is trained with multiple non-overlapping datasets. For Iter1 and Iter2, interactions between tasks are performed within iFAN. We can also observe that within iFAN, more iterations of interactions help the landmark localization achieve lower normalized mean error. Compared with the case of a single landmark localization task, the incorporation of the second task, face parsing, improves the performance of the baseline by about $2\%$, even though the face parsing dataset does not contain any duplicate image in the landmark localization dataset. With more iterations of task interactions between facial landmark localization and face parsing, the normalized mean error can be further decreased to $6.19\%$. We can see that multiple iterations of interactions between these two tasks gives rise to about $1.8\%$ improvement. The results clearly demonstrate that the iFAN model is powerful at exploiting the informative feedback during the task interactions, and the proposed cross-dataset hybrid learning is effective at learning useful knowledge from non-overlapping datasets with orthogonal annotations. 

\begin{figure}[t]
\includegraphics[width=\linewidth]{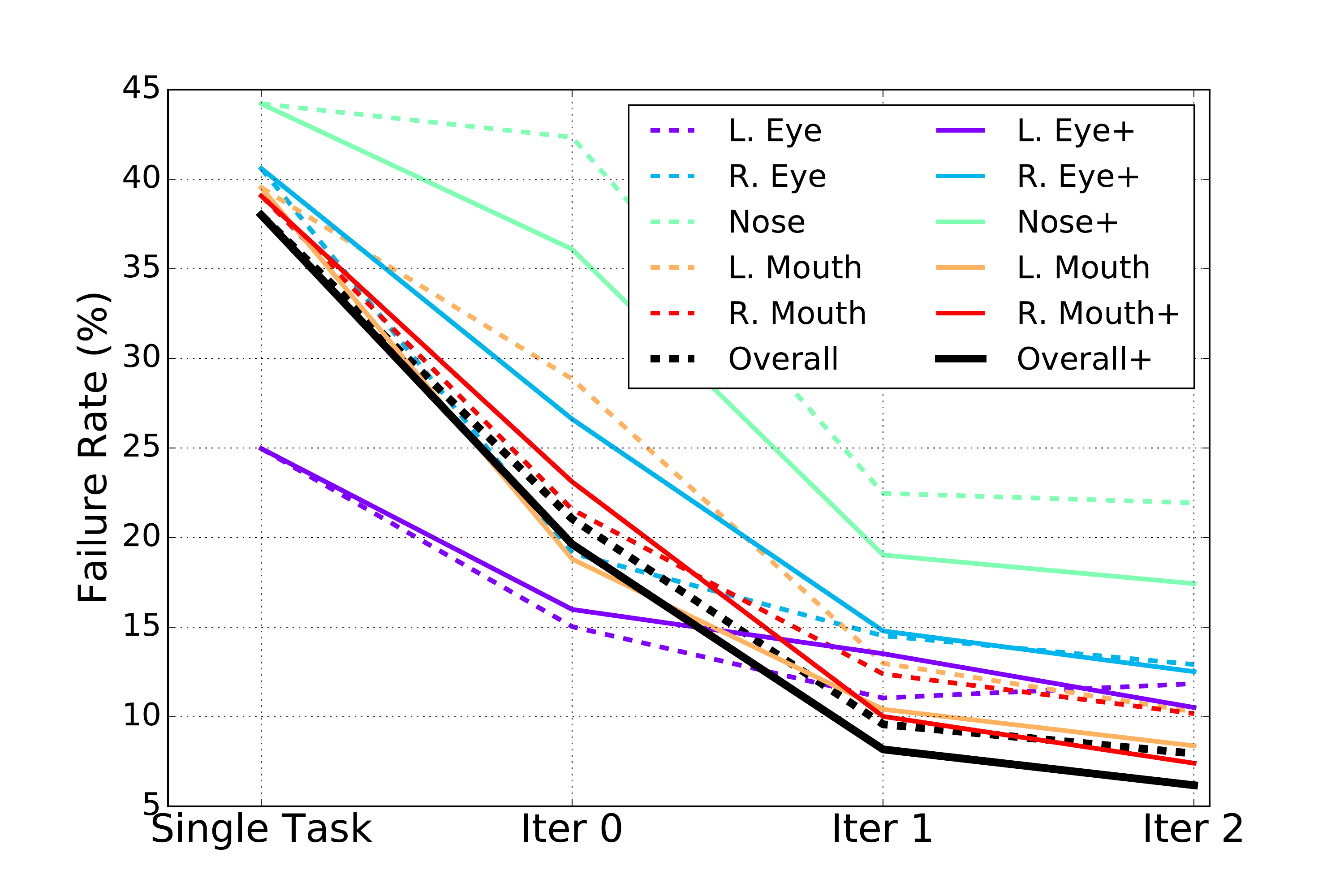}

\caption{Failure rate of facial landmark localization for different landmarks and different numbers of tasks. The dash lines denote the first setting (2T) and the solid lines denote the second setting (3T), which has a ``$+$'' mark on the corresponding legend.}\label{fig:pts_failure_rate}
\end{figure}

The proposed iFAN can also integrate $3$ different tasks into a single model and perform simultaneously well for all the $3$ tasks, as can be observed from the 3T cases. iFAN effectively exploits emotion information and provides informative cues (\emph{e.g.} movement of mouth corners) for the landmark localization task through the task integrator and feedback connections. The incorporation of the emotion recognition task helps improve the performance of landmark localization by about $0.35\%$.  The failure rates of different iterations corresponding to 2T and 3T cases are shown in Figure~\ref{fig:pts_failure_rate}. We can see that the trend is similar to Table~\ref{tab:pts}. Some qualitative examples from iFAN are shown in Figure~\ref{fig:quality}.

\begin{table}[b]\setlength{\tabcolsep}{2pt}

\caption{Normalized Mean Error (NME) (in $\%$) on MTFL dataset.  Different iterations of interactions with two and three tasks are shown.}\label{tab:pts}

\begin{tabular}{L{19mm}cccccc}
\toprule
& L.Eye & R.Eye & Nose & L.Mouth & R.Mouth & Mean  \\ 
\midrule
Single Task & 8.19 & 10.30 & 10.86 & 10.25 & 10.68 & 10.06 \\  
iFAN 2T Iter0& 6.52 & 8.21 & 9.67 & 7.39 & 8.03 & 7.96 \\  
iFAN 2T Iter1& 6.20 & 5.97 & 7.53 & 5.79 & 5.76 & 6.25 \\  
iFAN 2T Iter2& 5.99 & 6.10 & 7.46 & 5.73 & 5.68 & 6.19 \\  
\hline
iFAN 3T Iter0& 6.08 & 7.54 & 8.92 & 7.42 & 7.79 & 7.55 \\  
iFAN 3T Iter1& 5.93 & \textbf{5.91} & \textbf{6.79} & 5.38 & 5.26 & 5.85 \\  
iFAN 3T Iter2& \textbf{5.73} & 6.05 & 6.85 & \textbf{5.31} & \textbf{5.25} & \textbf{5.84} \\  
\bottomrule
\end{tabular}
\end{table}

\begin{table*}[!t]\setlength{\tabcolsep}{2pt}
\begin{center}
\caption{F-score (in $\%$) on Helen dataset for face parsing. 2T indicates there is another task jointly learned with the face parsing. 3T indicates there are in total three tasks. Note iFAN Iter0 corresponds to the standard multi-task learning.}\label{tab:seg}

\begin{tabular}{ lccccccccccc }
\toprule
& Eyes & Brows & Nose & In mouth & Upper Lip & Lower Lip & Mouth-All & Face Skin & Hair & Background & Overall \\ 
 \midrule
GSRM\cite{gu2008generative} & 74.3 & 68.1 & 88.9 & 54.5 & 56.8 &59.9 & 78.9 & - &- &- & 74.6 \\
Exemplar\cite{smith2013exemplar} & 78.5 &72.2 &92.2 &71.3 &65.1 &70.0 &85.7 &88.2 &- &- & 80.4 \\  
Multi-Objective\cite{liu2015multi} & 76.8 &73.4 &91.2 &82.4 &60.1& 68.4 &84.9& 91.2 &- &- & 85.4\\   
iCNN\cite{zhou2015interlinked} & \textbf{87.4} &81.3 &\textbf{95.0} &83.6 &75.4& 80.9 &92.6& - &- &- & 87.3\\   
\hline 
Single Task & 84.38 & 80.26 & 92.34 & 77.64 & 75.93 & 82.45 & 90.46 & 92.84 & 76.19 & 90.61 & 88.75 \\
iFAN 2T Iter0& 86.66 & \textbf{82.27} & 93.53 & 83.79 & 76.97 & 85.78 & 92.70 & 94.58 & 85.57 & 94.09 & 90.52\\
iFAN 2T Iter1& 86.60 & 82.22 & 94.03 & 85.62 & 78.87 & 87.13 & 93.79 & 94.68 & 85.90 & 94.05 & 91.03 \\
iFAN 2T Iter2& 86.59 & 82.20 & 94.07 & \textbf{86.63} & 79.25 & 87.48 & 93.98 & 94.67 & 85.91 & 94.04 & 91.10\\
\hline
iFAN 3T Iter0& 86.81 & 81.43 & 94.09 & 85.47 & 79.78 & 87.59 & 93.86 & \textbf{94.73} & \textbf{86.59} & \textbf{94.39} & 90.96 \\
iFAN 3T Iter1& 86.82 & 81.65 & 94.22 & 86.37 & 80.28 & 88.01 & 94.17 & 94.71 & 86.16 & 94.23 & 91.14 \\
iFAN 3T Iter2& 86.81 & 81.67 & 94.22 & \textbf{86.63} & \textbf{80.35} & \textbf{88.12} & \textbf{94.19} & 94.71 & 86.11 & 94.21 & \textbf{91.15} \\
\bottomrule
\end{tabular}
\end{center}
\end{table*}

\subsubsection{Face Parsing}
The performance on face parsing with iFAN and other baselines is listed in Table~\ref{tab:seg}. We can see that compared with other methods, iFAN achieves a new state-of-the-art performance in terms of overall F-score. Particularly, Multi-Objective~\cite{liu2015multi} formulates face parsing as a conditional random field with unary and pairwise classifiers and designs a multi-object learning method for this task. In contrary, in iFAN the face parsing task is only guided by the single unary classifiers,  and still outperforms Multi-Objective by a large margin. iCNN~\cite{zhou2015interlinked} consists of multiple CNNs taking input of different scales with an interlinking layer, which performs facial parts localization and pixel identification in a two-stage pipeline. In iFAN, only one singe model is used in an end-to-end network, which still outperforms iCNN by $4\%$ in terms of F-score. We can see that the strong baseline of fully convolutional DenseNet~\cite{jegou2016one} already outperforms iCNN in the Single Task case. Within iFAN, the incorporation of the facial landmark localization task improves the overall F-score of the face parsing task by about $2\%$ and the interactions between face parsing and facial landmark localization further improve the F-score by $0.6\%$ in the 2T case. So compared with iCNN, strong baseline architecture contributes to $1.5\%$ of performance gain, incorporation of facial landmark localization contributes to $2\%$ and the task interaction contributes to $0.6\%$. In the 3T case, iFAN gets slightly performance gain on face paring after the incorporation of the emotion recognition task. Some qualitative examples for face parsing from iFAN are shown in Figure~\ref{fig:quality}.

\subsubsection{Facial Emotion Recognition}
\begin{figure}
\includegraphics[width=0.49\linewidth]{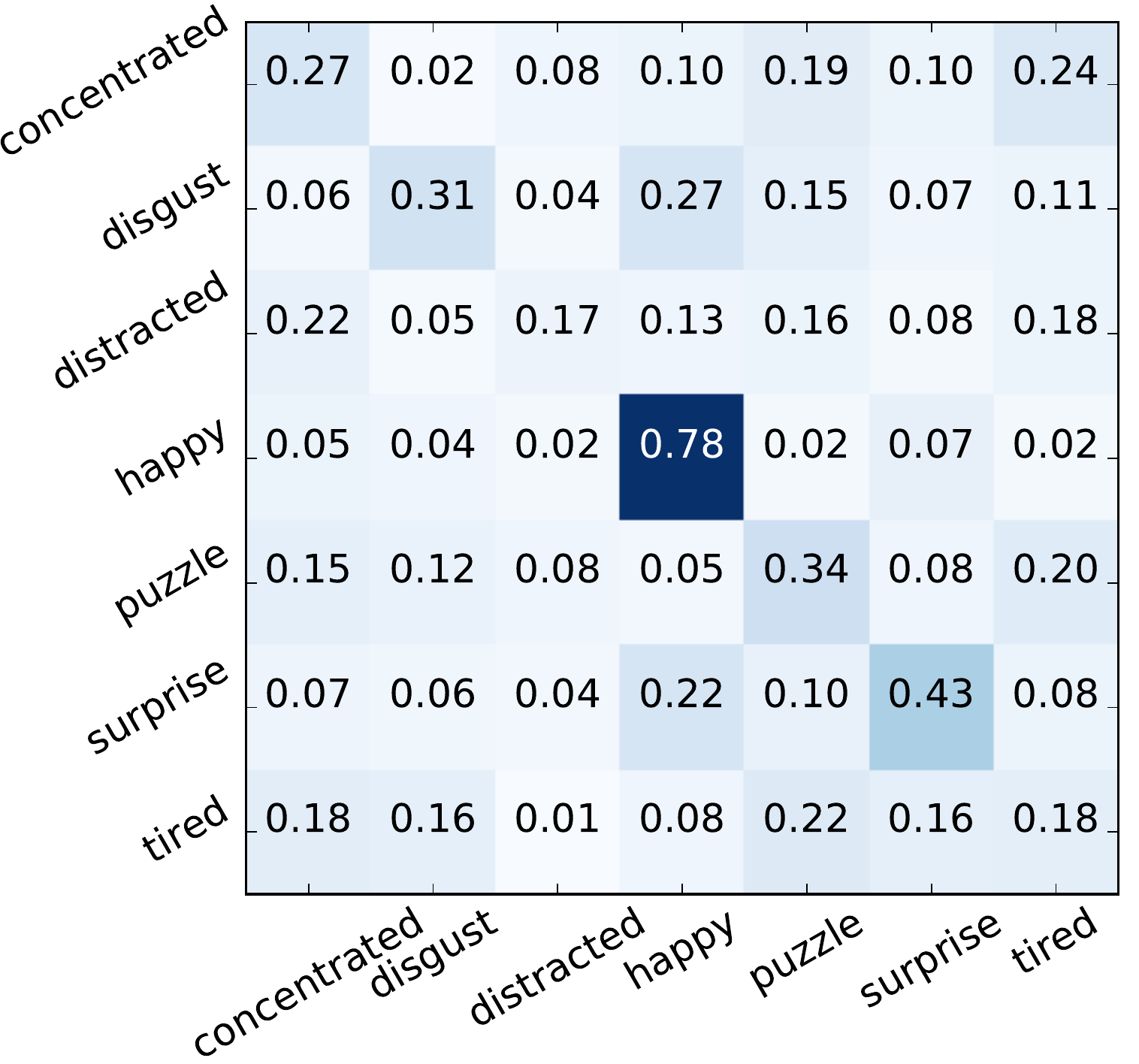}
\includegraphics[width=0.49\linewidth]{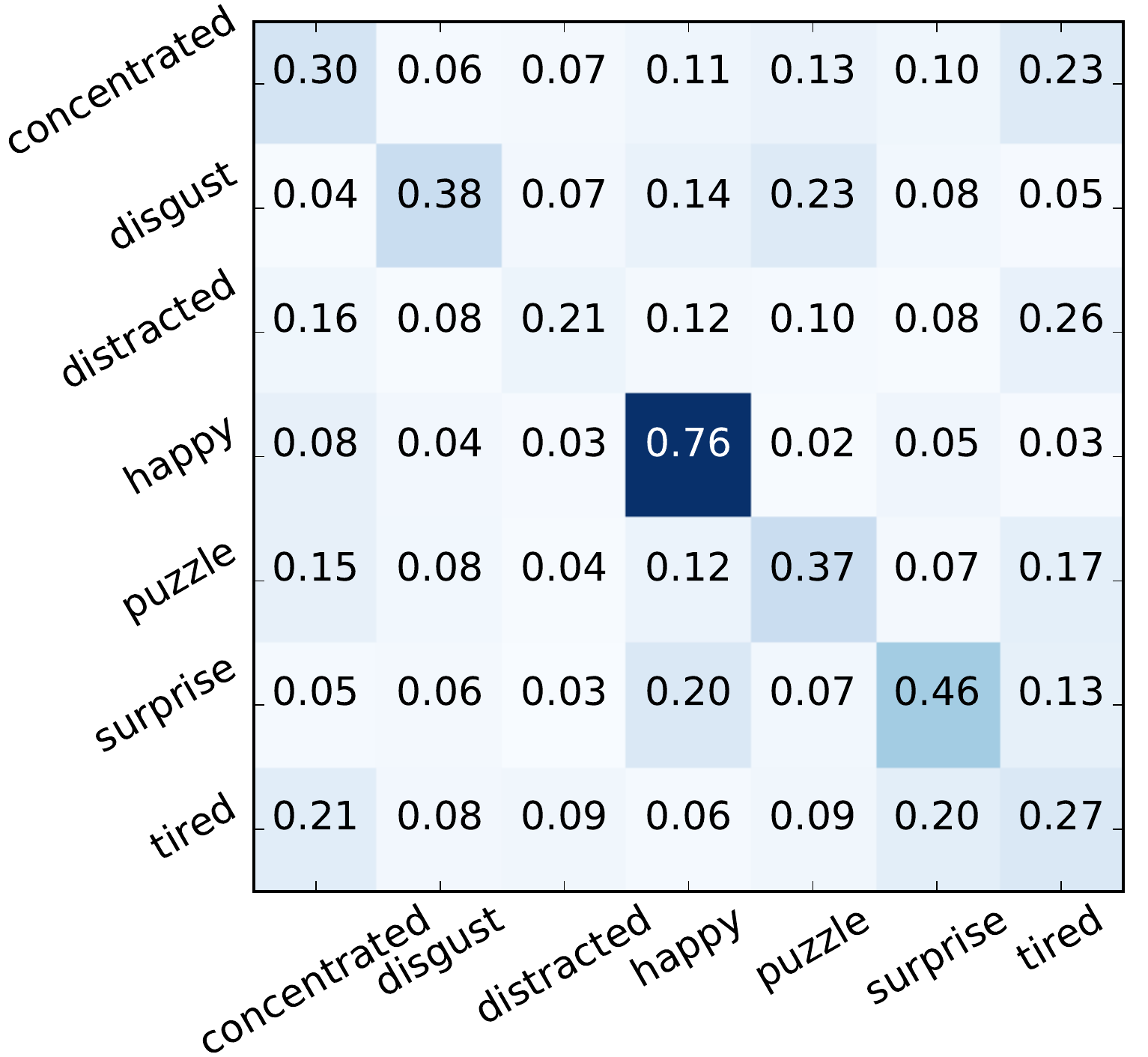}

\caption{The confusion matrices of different models. On the left panel is the output from the baseline model with no task interaction. On the right panel is the output from the proposed iFAN.}\label{fig:emo_confusion}

\end{figure}

For the facial emotion recognition task, we consider the following models: 1) a baseline model performing only emotion recognition on cropped faces; 2) a baseline model performing only emotion recognition on aligned faces; 3) iFAN performing three tasks simultaneously. The inputs to the integrated network are cropped faces. The performance on emotion recognition with different models is summarized in Table~\ref{tab:emo}. The confusion matrices corresponding to the first baseline model above and iFAN are shown in Figure~\ref{fig:emo_confusion}. While the traditional face alignment methods require facial landmark detection and face transformation (mapping the detected landmarks to some manually defined canonical locations) as pre-processing steps, we rely on the task interaction to perform alignment-free emotion recognition. We argue that by integration of the emotion recognition task with other related tasks (such as facial landmark localization), the emotion recognition task can be solved more effectively in iFAN than the traditional face alignment based pipeline. This is validated by the experimental results. Some qualitative examples for emotion recognition, together with the other two tasks are shown in Figure~\ref{fig:quality}. 

\begin{table}[!b]\setlength{\tabcolsep}{2pt}
\begin{center}

\caption{Facial emotion recognition accuracy of different models.}\label{tab:emo}
\begin{tabular}{lc}
\toprule
& Accuracy(\%)     \\ 
\midrule
Cropped Face& 42.26\\  
Aligned Face& 43.31\\
iFAN Iter0&  44.84\\  
iFAN Iter1&  45.16\\  
iFAN Iter2&  \textbf{45.40}\\  
\bottomrule
\end{tabular}
\end{center}

\end{table}

\subsection{Ablation Study}
We evaluate the effects of the two key components in our proposed iFAN, including the task integrator and the feature re-encoders, as well as the contribution of the cross-dataset hybrid training strategy to the final performance. 
\subsubsection{Task Integrator}
We have demonstrated the effectiveness of the task integrator on different tasks when it is not utilized and utilized for one or two times. To further probe the behavior of the task integrator with more iterations of task integrations, we perform additional iterations of interactions between tasks, and find that further iterations only provide marginal performance improvement as shown in Table~\ref{tab:ablation_feedback}.
The convergence is quickly achieved within one or two iterations of interactions.

\begin{table}\setlength{\tabcolsep}{2pt}

\begin{center}
\caption{Behavior of the task integrator with more iterations througth the feedback connections.}\label{tab:ablation_feedback}

\begin{tabular}{lccc}
\toprule
 & Overall F-score(\%) & NME(\%)  & Accuracy (\%)\\ 
\midrule
Single Task & 88.750 & 10.06 & 42.26\\  
iFAN Iter0& 90.961 & 7.55& 44.84\\  
iFAN Iter1& 91.142 & 5.85& 45.16\\
iFAN Iter2& \textbf{91.147} & 5.84& 45.40\\  
iFAN Iter3& 91.145 & \textbf{5.81}& \textbf{45.73}\\  
iFAN Iter4& 91.145 & 5.82& 45.48\\  
\bottomrule
\end{tabular}
\end{center}

\end{table}

\subsubsection{Feature Re-Encoders}
We then probe the effect of the feature re-encoders. We remove the feature re-encoders and replace them with simple resizing operation to directly convert the prediction maps (\emph{i.e.} the input into the feature re-encoders) to the size of the respective feature map for the purpose of task interaction. In this way, the predictions of different tasks are used in their original feature space and no encoding is performed. We find that the normalized mean error of landmark localization increases to $10.5\%$, the accuracy of the emotion recognition drops to $42.09\%$ and the F-score of face parsing drops to $89.4\%$ after two iterations of interactions. We can see that the feature re-encoders facilitate better interactions between different tasks. 
 
\subsubsection{Cross-dataset Hybrid Training Strategy}
In the cross-dataset hybrid training strategy,  task dependent batch normalization parameters are used. When we enforce all the tasks to share the same batch normalization parameters, the performance after two iterations reduces to $9.74\%$, $33.63\%$ ad $87.65\%$ for facial landmark localization, facial emotion recognition and face parsing, respectively. We can see that task-wise batch normalization parameters give rise to remarkable performance boost in the proposed iFAN.

\begin{figure}
\includegraphics[width=0.9\linewidth]{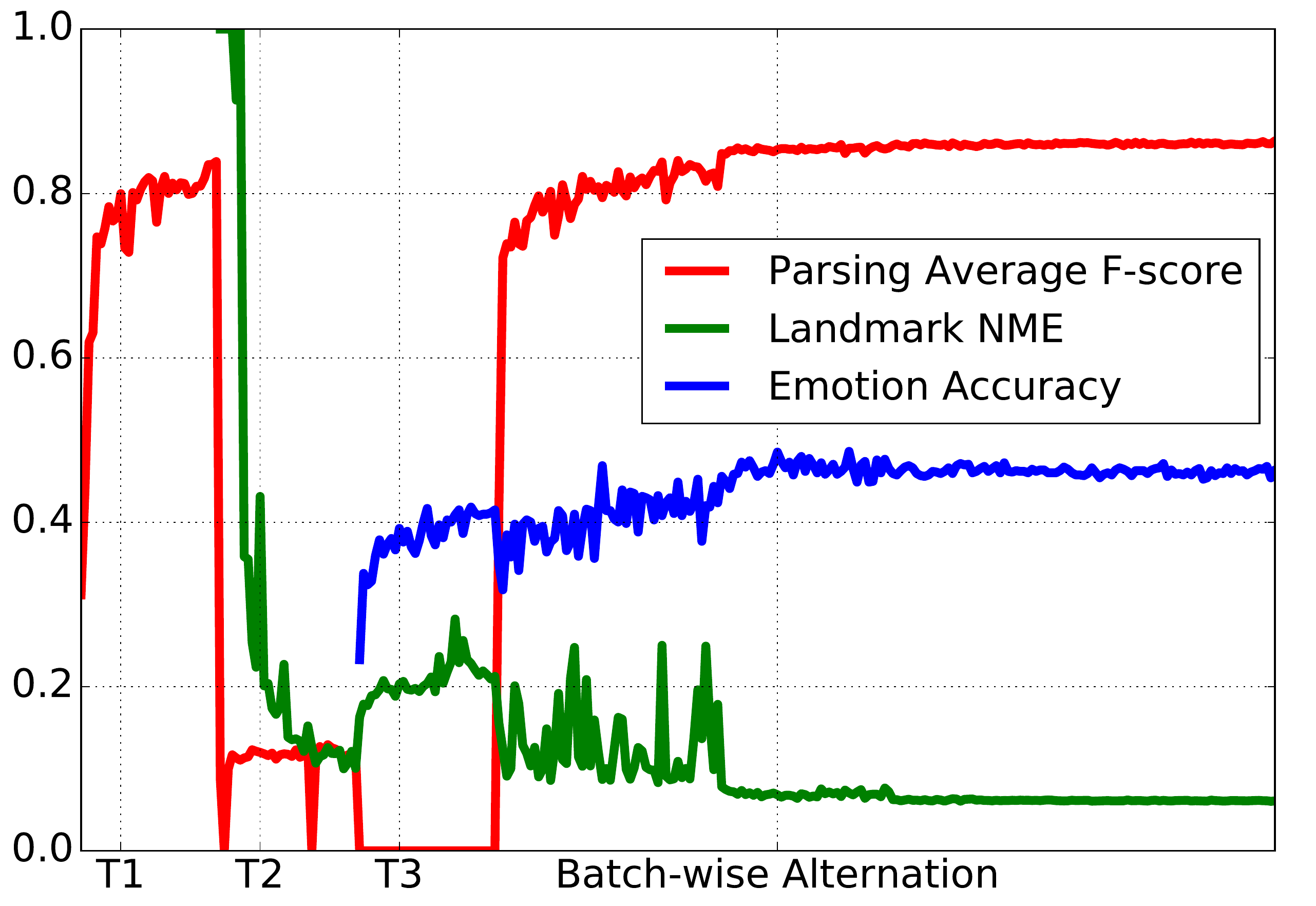}

\caption{Performance of three tasks in different stages of the cross-dataset hybrid training strategy. Note that for landmark detection (the green curve), lower numbers mean better performance.}
\label{fig:alternative}

\end{figure}

There are two stages in the cross-dataset hybrid training strategy: task-wise pre-training and batch-wise fine-tuning. For the task wise pre-training, the training of one task will negatively affect performance of other tasks. To illustrate the process, the metrics of three tasks in different stages of the optimization process are shown in Figure~\ref{fig:alternative}. T1 denotes the pre-training stage of the first task (face parsing), where the parsing average F-score is increasing. We can see during the pre-training of the second task (facial landmark), denoted by T2, the performance of facial landmark localization is increasing (lower normalized mean error), but the performance of parsing is decreasing quickly. During the pre-training of the third task, we can observe performance decreasing for both the first two tasks. The reason is that different tasks are trained on different datasets and the network easily biases to one of them during the pre-training stage. In the batch-wise alternative fine-tuning stage, we can see the performance of all the three tasks is increasing. With the batch-wise alternative fine-tuning, the performance can gradually get back to that of the pre-training stage, and then it is further improved through task interactions.

\section{Conclusion}

In this work, we proposed an integrated face analytics network iFAN that performs multiple face analytics tasks simultaneously. The proposed iFAN fully exploits the correlations between tasks and enables interactions between them. The feature re-encoders and task integrator in iFAN facilitate better task interactions and integrations. With the  cross-dataset hybrid training strategy, the proposed network is able to learn from multiple data sources with no overlapping labels, allowing the ``plug-in and play'' feature for practical usage in multimedia applications.

\section*{Acknowledgement}
This work was partially funded by National Research Foundation of Singapore. The work of Jiashi Feng was partially supported by NUS startup R-263-000-C08-133, MOE Tier-I R-263-000-C21-112 and IDS R-263-000-C67-646.

\newpage
\bibliographystyle{ACM-Reference-Format}
\bibliography{face} 

\end{document}